\documentclass[sigconf]{acmart}

\AtBeginDocument{%
  }

\setcopyright{acmcopyright}
\copyrightyear{2018}
\acmYear{2018}
\acmDOI{XXXXXXX.XXXXXXX}

\acmConference[Conference acronym 'XX]{Make sure to enter the correct
  conference title from your rights confirmation emai}{June 03--05,
  2018}{Woodstock, NY}

\acmSubmissionID{82}

\usepackage{graphicx}
\usepackage{subcaption}
\usepackage{caption}

\usepackage{cleveref}
\usepackage{float}
\usepackage{lscape}                                         

\usepackage[lined,ruled,linesnumbered]{algorithm2e}

\usepackage{booktabs}                   
\usepackage{multirow}

\usepackage{paralist}
\usepackage{enumitem}

\usepackage{bm}                          
\usepackage{epsfig}                      
\usepackage{amsmath}   
\usepackage{pifont}

\usepackage{units}
\usepackage{color}

\usepackage{comment}

\usepackage{url}  
\usepackage{xspace}
\usepackage{setspace}
\usepackage[percent]{overpic}
\usepackage{nccbbb}
\usepackage{ifthen}

\def\eg{e.g.,~}               

\newlength\paramargin
\newlength\figmargin
\newlength\secmargin
\newlength\figcapmargin

\newcommand{\heading}[1]
{
\vspace{1mm}
\noindent \textbf{#1}
}   

\newcommand{\secref}[1]{Section~\ref{sec:#1}}
\newcommand{\figref}[1]{Figure~\ref{fig:#1}} 
\newcommand{\tabref}[1]{Table~\ref{tab:#1}}
\newcommand{\eqnref}[1]{Equation~\ref{eq:#1}}

\long\def\ignorethis#1{}

\newcommand{\tb}[1]{\textbf{#1}}

\graphicspath{{figure}, {images}, {example}}

\def\styleimg{I}

\def\semanticmapset{\mathbb{S}}
\def\semanticmap{s_i}
\def\wstyle{\mathrm{w}^{sty}}
\def\wgeo{\mathrm{w}^{struct}}

\def\wfull{\mathrm{w}^{full}}

\def\wspace{\mathrm{w}}
\def\wplusspace{\mathcal{W}+}
\def\estyle{E_{sty}}

\def\egeo{E_{struct}}

\def\styleGAN{G}
\def\networkName{CTGAN}

\def\texturecolorset{\mathbb{T}}
\def\texturecolor{t_i}
\def\shapeinput{M}
\def\texturedmodel{R}
\def\uvmapset{\mathbb{U}}
\def\uvmap{u_i}

\begin{document}

\title{\networkName: Semantic-guided Conditional Texture Generator for 3D Shapes}

\author{Yi-Ting Pan}
\affiliation{%
  \institution{National Taiwan University}
  \country{Taiwan}
}

\author{Chai-Rong Lee}
\affiliation{%
  \institution{National Tsing Hua University}
  \country{Taiwan}
}

\author{Shu-Ho Fan}
\affiliation{%
  \institution{National Tsing Hua University}
  \country{Taiwan}
}

\author{Jheng-Wei Su}
\affiliation{%
  \institution{National Tsing Hua University}
  \country{Taiwan}
}

\author{Jia-Bin Huang}
\affiliation{%
  \institution{University of Maryland College Park}
  \country{USA}
}

\author{Yung-Yu Chuang}
\affiliation{%
  \institution{National Taiwan University}
  \country{Taiwan}
}

\author{Hung-Kuo Chu}
\affiliation{%
  \institution{National Tsing Hua University}
  \country{Taiwan}
}
\email{hkchu@cs.nthu.edu.tw}



\begin{abstract}
The entertainment industry relies on 3D visual content to create immersive experiences, but traditional methods for creating textured 3D models can be time-consuming and subjective. Generative networks such as StyleGAN have advanced image synthesis, but generating 3D objects with high-fidelity textures is still not well explored, and existing methods have limitations.
We propose the Semantic-guided Conditional Texture Generator (\networkName), producing high-quality textures for 3D shapes that are consistent with the viewing angle while respecting shape semantics. {\networkName} utilizes the disentangled nature of StyleGAN to finely manipulate the input latent codes, enabling explicit control over both the style and structure of the generated textures. A coarse-to-fine encoder architecture is introduced to enhance control over the structure of the resulting textures via input segmentation.
Experimental results show that {\networkName} outperforms existing methods on multiple quality metrics and achieves state-of-the-art performance on texture generation in both conditional and unconditional settings. We will release our code and data after publication.


\end{abstract}


\begin{teaserfigure}
  \includegraphics[width=\textwidth]{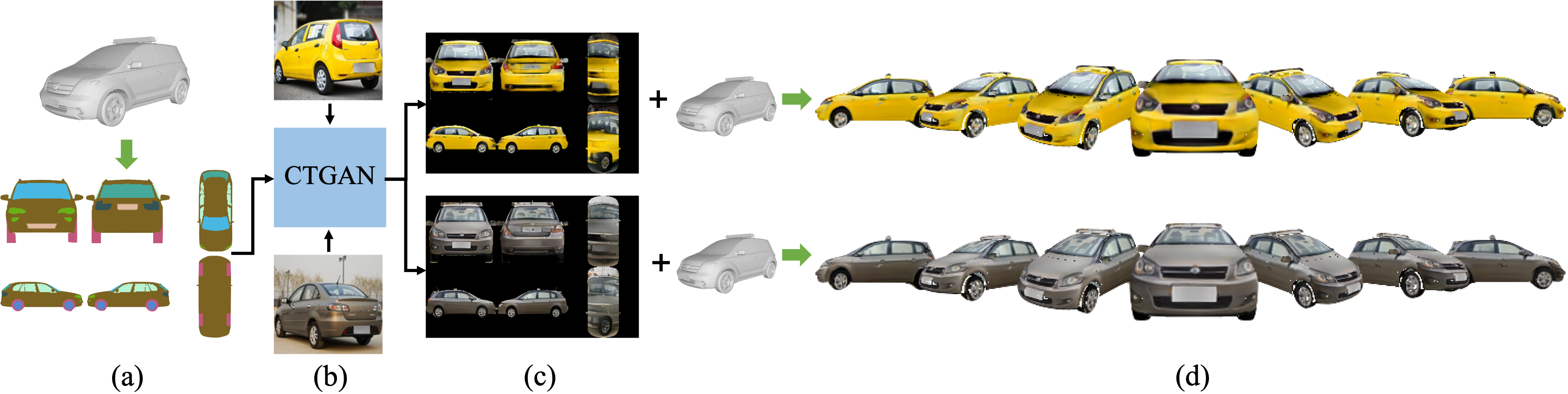}
  \caption{\tb{Conditional texture generation}. We propose a new conditional texture generator (CTGAN) that is conditioned on the segmentation map derived from the 3D model (a) and style images (b) and is able to generate high-quality texture maps (c) for 3D objects. Our textured 3D models (d) are view consistent and aligned with the input style images.}
  \Description{}
  \label{fig:teaser}
\end{teaserfigure}


\maketitle

\section{Introduction}
\label{sec:introduction}





3D visual content has become increasingly essential in entertainment, especially in movies, games, and virtual reality applications. These fields rely on 3D models to create immersive and engaging user experiences. 
The most common representation of 3D content includes both geometry and texture information. 
Traditional methods for creating textured 3D models often require multiple steps and can be time-consuming for professionals. 
Manually creating textures involves a skilled artist hand-painting textures on the 3D model or capturing photographs of real-world objects to apply them as textures. These traditional methods can be subjective and labor-intensive and may not always result in the desired level of realism. 

Advancements in generative learning techniques and the availability of large-scale 3D shape datasets have significantly enhanced the modeling of 3D geometry, resulting in highly realistic geometric representations of 3D objects. However, generating 3D objects with high-fidelity textures remains an area that has not been extensively explored.
Meanwhile, Generative Adversarial Networks (GANs), especially the style-based GAN architecture known as StyleGAN~\cite{karras2019style,karras2020analyzing,Karras2020ada}, have significantly advanced image synthesis, achieving high visual quality and generating images with remarkable realism. Previous research~\cite{yang2019semantic,shen2020interpreting,Collins20} has also shown that StyleGAN has a disentangled latent code that enables control and editing of the generated results.

Recent research~\cite{rui2021LTG,siddiqui2022texturify} has aimed to learn the texture generator from real-world images by utilizing the generative capabilities of StyleGAN~\cite{karras2019style,karras2020analyzing}. These methods have produced superior visual quality when compared to learning implicit fields to model textures in 3D space~\cite{OechsleICCV2019}, as demonstrated in~\figref{banner_compare}. However, existing methods have several limitations. Firstly, these models were designed to generate textures from a latent code randomly sampled from the latent space, lacking direct control to generate textures with style conditioned on a given reference image. Secondly, using a separated StyleGAN~\cite{karras2019style} generator for generating textures at different views may result in inconsistent style after texture mapping to the 3D shapes, as illustrated in~\figref{banner_compare}. Lastly, the generated texture may not align with the underlying shape semantics due to either weak supervision on the shape silhouette~\cite{rui2021LTG} or insufficient mesh resolution~\cite{siddiqui2022texturify}.

To overcome the aforementioned limitations, we propose a new approach named the \emph{Semantic-guided Conditional Texture Generator} (\networkName), which can produce high-quality textures for 3D shapes consistent with the viewing angle. As shown in~\figref{teaser}, our approach enables controllable texture generation while respecting the shape semantics.
Our method is based on the LTG architecture~\cite{rui2021LTG}, which employs a canonical-view projection-based texture atlas for texture parameterization and uses StyleGAN to generate textures for each canonical view. However, instead of using a separate StyleGAN generator for each texture map, we utilize the StyleGAN2-ADA~\cite{Karras2020ada} which inherits the disentangled nature of StyleGAN~\cite{karras2019style} to finely manipulate the input latent codes, enabling explicit control over the style and structure of the generated textures and achieving more stable training~\cite{Karras2020ada} than StyleGAN~\cite{karras2019style} on small dataset.
We separate the disentangled latent codes of StyleGAN2-ADA~\cite{Karras2020ada} into two parts: the \emph{structure} codes and the \emph{style} codes, which control the structure and appearance of the resulting textures, respectively. We then train two encoders, the structure encoder, and style encoder, to map a semantics segmentation map and a reference style image into their respective latent codes. We employ a coarse-to-fine encoder architecture for the structure encoder to represent the input semantic map better. 
We evaluate our approach on the ShapeNet car dataset~\cite{shapenet2015} and the FFHQ human face dataset~\cite{Karras_2019_CVPR}, and compare it with state-of-the-art texture generation methods. The experimental results demonstrate that our approach achieves a significant performance boost in both conditional (FID score drops from 139.63 to 39.41) and unconditional (FID score drops from 87.06 to 66.73) settings.
In summary, we make the following contributions:
 

\begin{figure}[t]
    \centering
    \includegraphics[width=.49\textwidth]{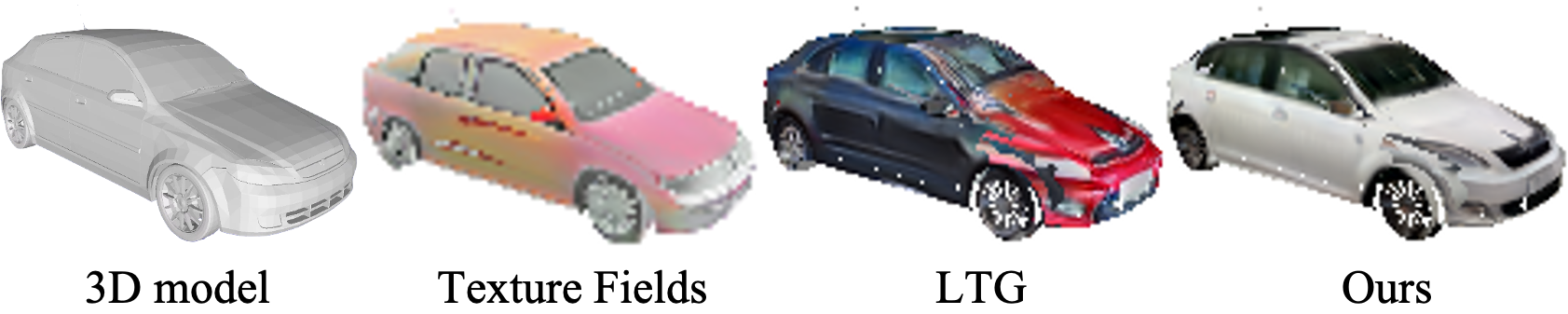}
    \caption{\tb{Limitation of existing methods.} 3D model texture generation by existing methods suffer from producing unrealistic and blurry results (Texture Fields~\cite{OechsleICCV2019}), and generating view-inconsistent texture (LTG~\cite{rui2021LTG}).}
    
    \label{fig:banner_compare}
\end{figure}

\begin{itemize}
    \item We propose a novel Semantic-guided Conditional Texture Generator (\networkName), that effectively utilizes the disentangled nature of StyleGAN2-ADA to enable explicit control over both the style and structure of the generated textures, resulting in high-quality texture-mapped 3D shapes.
    %

    \item We introduce a coarse-to-fine encoder architecture to enhance control over the structure of the resulting textures via the input semantic segmentation.
    
    \item We achieve state-of-the-art performance on both the ShapeNet car and FFHQ human face datasets under conditional and unconditional settings compared to baselines.
\end{itemize}

\section{Related Work}
\label{sec:RelatedWork}
\begin{figure*}[t]
    \begin{overpic}[width=\textwidth]{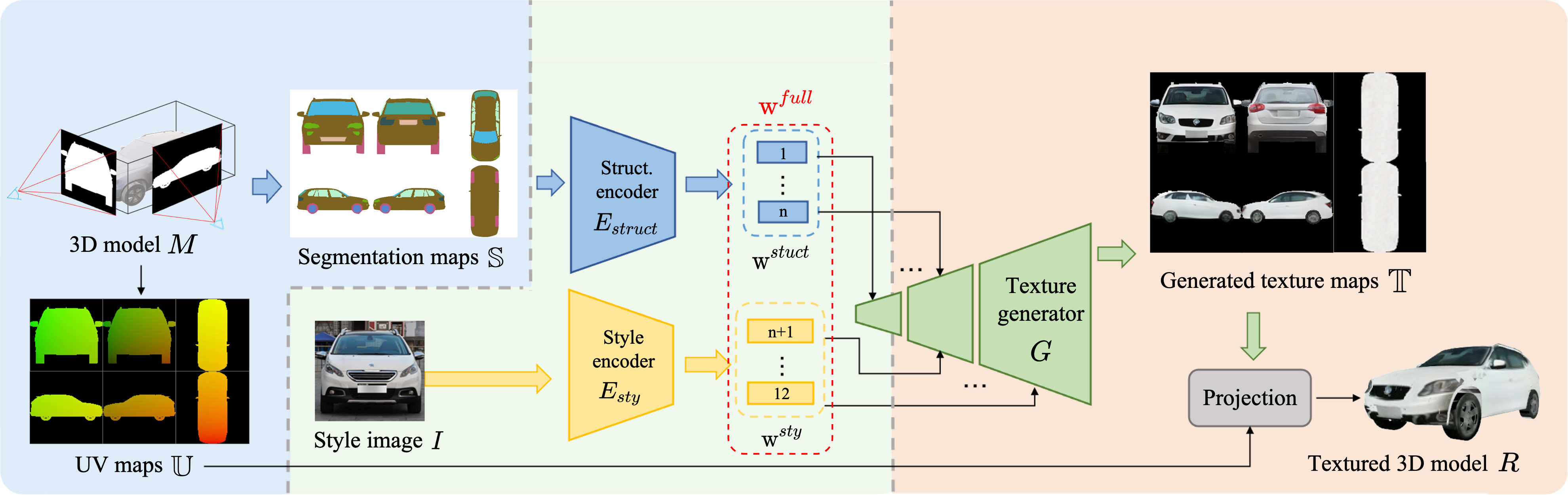}
    \put(0, 29){\begin{minipage}{0.35\textwidth}\centering Texture Parameterization  (\secref{textureMapping}) \end{minipage}}
    \put(33, 27.7){\begin{minipage}{0.25\textwidth}\centering Style Code Encoders \\(\secref{styleganEncoders}) \end{minipage}}
    \put(63, 29){\begin{minipage}{0.35\textwidth}\centering Texture Generator (\secref{textureGenerator}) \end{minipage}}
    \end{overpic}
    \caption{\tb{System overview.} Given 3D model $\shapeinput$ as input, we start with texture parameterization to generate the corresponding UV maps $\uvmapset$ and the segmentation maps $\semanticmapset$. 
    The texture generator $\styleGAN$ then takes style code $\wfull$ as input and generates the texture maps $\texturecolorset$ based on the segmentation maps $\semanticmapset$. 
    To ensure view-consistent results, we divide the style code $\wfull$ and separately encode segmentation maps $\semanticmapset$ and style image $\styleimg$ into the structure representation $\wgeo$ and the style representation $\wstyle$  using structure encoder $\egeo$ and style encoder $\estyle$.
    Finally, we apply our generated texture maps $\texturecolorset$ on the 3D model $\shapeinput$ and produce the 3D textured model $\texturedmodel$.}
    %
    \label{fig:pipeline}
\end{figure*}

\paragraph{Generative Model on 3D Domain.}
%
Generating 3D models with realistic appearance and texture has become popular in recent years. Early approaches relied on latent representations from the geometry and texture of the model~\cite{piGAN2021,OechsleICCV2019} or 3D disentangled representations~\cite{Nguyen-Phuoc_2019_ICCV,Niemeyer2020GIRAFFE}. 
Style-based GANs~\cite{karras2019style,karras2020analyzing} have achieved high-fidelity results by using disentangled latent codes to represent implicit attributes. Some works have introduced coordinate systems~\cite{NEURIPS2021_076ccd93} or extracted segmentation maps from implicit fields~\cite{sofgan,shi2021SemanticStyleGAN} to control 3D information, while others have learned 3D models from synthesized results\cite{gu2022stylenerf}. Another approach generates high-quality 2D images from the 3D model using disentangled latent codes~\cite{Chan2022,zhou2021CIPS3D,orel2022styleSDF}. 
However, these works focus on generating 2D results rather than 3D textured models. In contrast, our method aims to produce 3D shapes with a realistic texture that are conditioned on the given style images.

\paragraph{Latent Space Manipulation.}
To enable precise control over the output of GAN~\cite{goodfellow2014generative}, various methods for style latent code editing~\cite{Chiu2020HumanintheloopDS,shen2020interpreting,shen2021closedform} have been proposed. These approaches encode a given style image into a latent space and then edit the latent code in a meaningful manner. 
While existing methods are effective, using the segmentation maps to manipulate the latent code is still a less explored challenge.
To this end, we propose an architecture that combines both of them.
In detail, we use two encoders, structure encoder and style encoder, to encode both of the segmentation maps and style images into latent space and control the latent code with the segmentation maps and style images.



\paragraph{Texture Generation for 3D Shapes.}
In recent years, the generation of textures for 3D shapes has received considerable attention from researchers, as evidenced by several recent works~\cite{OechsleICCV2019, rui2021LTG, siddiqui2022texturify,gao2022get3d}. 
Texture Fields~\cite{OechsleICCV2019} utilizes implicit fields to model textures in 3D space, while LTG~\cite{rui2021LTG} uses a canonical-view projection-based texture atlas representation and employs a separate StyleGAN~\cite{karras2019style} to generate a texture map for each view. It also addresses texture inconsistency across views using multiple discriminators. Texturify~\cite{siddiqui2022texturify} generates textures directly on the surfaces of 3D shapes using convolutional operators that are conditioned on both 3D models and style latent codes. However, these methods have several limitations, including producing blurry results~\cite{OechsleICCV2019}, generating inconsistent textures across views~\cite{rui2021LTG}, and producing textured models of mediocre quality due to insufficient structure guidance~\cite{rui2021LTG} or low mesh resolution~\cite{siddiqui2022texturify} during the inference process. 

Some of the texture generation works focus on synthesizing pure texture without input mesh~\cite{Fruehstueck2019TileGAN,rodriguezpardo2022SeamlessGAN}, however, these methods cannot be used in our scenario since they do not provide UV maps.

Recently, the diffusion model~\cite{rombach2022high} has become popular. There are some works that combine the power of a text encoder~\cite{alec2021clip} with the diffusion model to synthesize textures given an input mesh~\cite{metzer2023latent,richardson2023texture,chen2023text2tex}.
However, text-prompt guided methods are not ideal for generating textures that require fine-grained control over semantics and structure.

%
In contrast, our method leverages the disentanglement nature of StyleGAN2-ADA~\cite{Karras2020ada} to explicitly control the structure and style of the generated textures through separated latent codes, enabling the conditional setting where our model learns the mapping from input structure and style maps using different encoders. As a result, we can generate high-quality textured 3D shapes that surpass those generated by prior methods.




\section{Overview}
\label{sec:overview}

\figref{pipeline} illustrates the system overview of {\networkName}, which consists of three blocks.
In the \emph{texture parameterization} block, we take the 3D model $\shapeinput$ as input, generate the corresponding 2D UV maps $\uvmapset= \{ \uvmap \}_{i=1}^N$, and label the segmentation maps $\semanticmapset= \{ \semanticmap \}_{i=1}^N$ aligned with UV maps $\uvmapset$ where $N$ is the number of view (\secref{textureMapping}).
As for texture generator block, we adopt StyleGAN2~\cite{Karras2020ada} as our texture generator $\styleGAN$ to generate the texture maps $\texturecolorset= \{ \texturecolor \}_{i=1}^N$ based on the segmentation maps $\semanticmapset$ (\secref{textureGenerator}).
In order to produce view-consistent results, we separate the style code $\wfull$ of StyleGAN2~\cite{Karras2020ada} into structure representation $\wgeo$ and style representation $\wstyle$ encoded from segmentation maps $\semanticmapset$ and style image $\styleimg$ using two encoders denoted as $\egeo$ and $\estyle$ respectively (\secref{styleganEncoders}).
Finally, we produce a 3D textured model $\texturedmodel$ by projecting our estimated texture maps $\texturecolorset$ back to the 3D model $\shapeinput$. We adopt a three-stage training process, first training the texture generator $\styleGAN$, followed by the style encoder $\estyle$, and finally the structure encoder $\egeo$ (\secref{lossfunction}).

\section{Method}
\label{sec:method}
\figref{pipeline} depicts the system overview of {\networkName}, which comprises three main components: texture parameterization, texture generator, and style code encoders. In the following subsections, we will provide a detailed explanation of each of these components.

\subsection{Texture Parameterization}
\label{sec:textureMapping}

To achieve our objective of generating textures for a particular 3D model, it is necessary to define the texture parameterization process for mapping 2D textures onto the 3D domain. We employ the canonical-view projection-based texture atlas, similar to LTG~\cite{rui2021LTG}, to parameterize the input 3D model with six UV maps. 
The process involves projecting the input 3D model onto $N$ predefined viewpoints, assigning the projected coordinate as a vertex's texture coordinate, and creating the corresponding UV maps $\uvmapset= \{ \uvmap \}_{i=1}^N$. Specifically, for the car shapes depicted in~\figref{pipeline}, we utilize six viewpoints, including the front, back, left, right, top, and bottom viewpoint.
We further generate the semantic segmentation maps $\semanticmapset= \{ \semanticmap \}_{i=1}^N$ for the projected views by manually defining the semantic parts or employing other off-the-shelf methods~\cite{zllrunning2019, zhang2021datasetgan} on the 2D projection of the input 3D model. These semantic segmentation maps serve as structural guidance for the subsequent texture generator.


%
%
%
%

\subsection{Texture Generator}
\label{sec:textureGenerator}

In this study, we employ the StyleGAN2-ADA~\cite{Karras2020ada} architecture as our texture generator to produce high-fidelity textures. The texture generator $\styleGAN$ accepts 12 1D vectors as input style codes, with each code corresponding to a block of the model. The generator outputs 2D texture maps $\texturecolorset$ in the projected views. To obtain the final textured 3D model, we map the 2D texture maps $\texturecolorset$ onto the 3D shape via the UV maps $\uvmapset$.
As mentioned earlier, StyleGAN~\cite{karras2019style} incorporates a disentangled latent code $\wspace \in \mathcal{W}$, which can be extended to a latent space $\wplusspace$ to provide better control and editing capabilities for the results. Various GAN inversion works~\cite{9008515, gansteerability, tov2021designing, 9578038} have demonstrated that $\wplusspace$ is capable of reconstructing better results compared to $\wspace$. Thus, we encode all style codes in $\wplusspace$.
To analyze which layers of style codes control structural attributes, such as pose, shape, semantic parts, etc., and what other layers control the style attributes, we adopt the StyleSpace~\cite{zongze2020stylespace} to separate the style codes $\wfull$ into two parts: the first $n$ layers control structural attributes, denoted as $\wgeo$, while the remaining layers control style attributes, denoted as $\wstyle$. This explicit separation in the latent code space enables our approach to support conditional texture generation, generate textures with better style consistency, and respect the given shape semantic guidance.

\begin{figure}[!t]
\centering
    \begin{subfigure}[t]{.4\columnwidth}
       \includegraphics[width=\columnwidth,keepaspectratio]{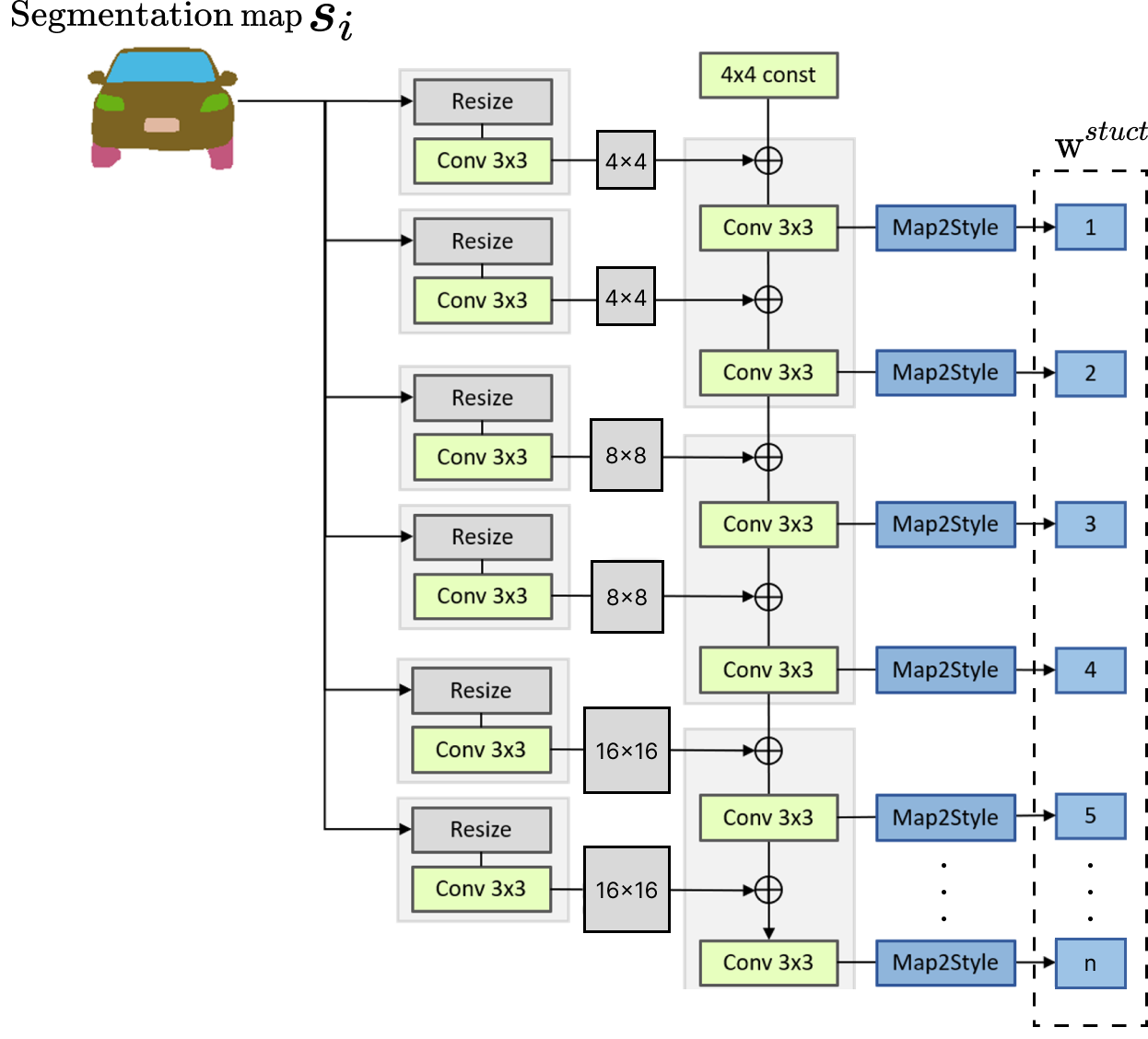}
       \caption*{(a) Coarse-to-fine structure encoder.}
       \label{fig:geo_encoders}
    \end{subfigure}
    \begin{subfigure}[t]{.58\columnwidth}
       \includegraphics[width=\columnwidth,keepaspectratio]{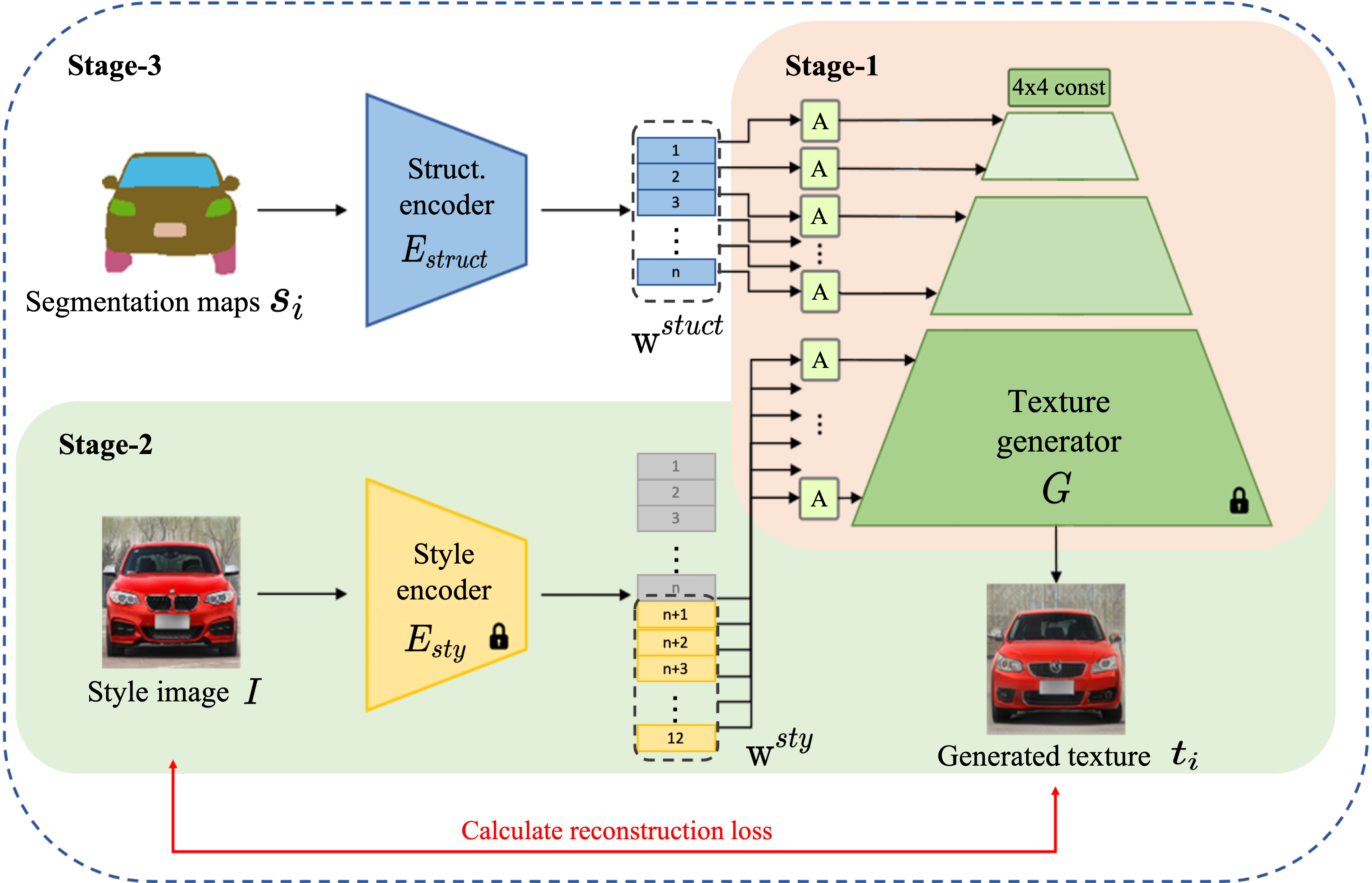}
       \caption*{(b) Training process.}
       \label{fig:training_process}
    \end{subfigure}
    \caption{(a) Coarse-to-fine structure encoder. Our coarse-to-fine structure encoder encodes segmentation maps $\semanticmap$ into structure latent code $\wgeo$ gradually from low (coarse) to high (fine) resolution.
    (b) We start by training the texture generator $\styleGAN$, then the style encoder $\estyle$, and finally the structure encoder $\egeo$.
    }
    \label{fig:two_figures}
\end{figure}

\subsection{Style Code Encoders}
\label{sec:styleganEncoders}


To explicitly control the structural and style attributes of the generated textures, we utilize two encoders: the \emph{structure encoder} $\egeo$ and the \emph{style encoder} $\estyle$. These encoders respectively map the input semantic segmentation map and style image into structural latent codes $\wgeo$ and style latent codes $\wstyle$. We adopt the pSp~\cite{richardson2021encoding} encoder as our style encoder $\estyle$ and take the $n+1^{th}$ to $12^{th}$ layers of the output of the \emph{style encoder} $\estyle$ as the style latent codes $\wstyle$.
As the semantic segmentation map contains structural information of different levels of detail, we devise a coarse-to-fine architecture, as depicted in~\figref{two_figures}~(a), to better preserve the semantic information, as suggested by previous works~\cite{Zhu_2020_CVPR,park2019SPADE,kim2021stylemapgan,gunhee2022stylelangan}.
Specifically, the structure encoder $\egeo$ resizes the input segmentation maps $\semanticmap$ to the same size as the feature map in each layer of the texture generator $\styleGAN$. The resized segmentation maps are then fed into the convolution layers, and the output feature maps are added to the results of the previous layer, before the feature maps are transformed into 1D vectors through the Map2Style~\cite{richardson2021encoding} block. The structure latent codes $\wgeo$ are formed by all 1D vectors encoded by the structure encoder $\egeo$.

\subsection{Loss Function and Training Process}
\label{sec:lossfunction}

As illustrated in~\figref{two_figures}~(b), we adopt a 3-stage training process to train our network. 
First, we train the texture generator $\styleGAN$ using a collection of real-world images following the same training procedure and the loss function of the StyleGAN2-ADA~\cite{Karras2020ada}.
%
Second, we train the style encoder $\estyle$ using a collection of real-world images as the training data. Since the goal of style encoder $\estyle$ is to find style codes that correspond to the style image, we thus calculate the reconstruction loss between the input style image $\styleimg$ and the generated texture $\texturecolor$.
Finally, we train the structure encoder $\egeo$ using the segmentation maps of the same collection of real-world images as the training data. Since the input segmentation map $\semanticmap$ is derived from the input style image $\styleimg$, we adopt the similar reconstruction loss between the input style image $\styleimg$ and the generated image $\texturecolor$ to train the structure encoder $\egeo$ while the texture generator $\styleGAN$ and the style encoder $\estyle$ are both frozen.

Following the approach proposed in \cite{richardson2021encoding}, we used a weighted combination of several objectives as the reconstruction loss function to train the two encoders. The first objective is the pixel-wise $\ell_2$ loss, which is defined as:
\begin{equation}
    \ell_{2}( \styleimg, \texturecolor ) = || \styleimg - \texturecolor ||_{2}
\end{equation}
where $\styleimg$ is the original image and $\texturecolor$ is the synthesized image.

To learn perceptual similarities, we also utilized the LPIPS~\cite{zhang2018lpips} loss, which measures the distance between the feature representations of the two images, and is defined as:
\begin{equation}
    \ell_{LPIPS}( \styleimg, \texturecolor ) = ||F(\styleimg) - F(\texturecolor)||_{2}
\end{equation}
where ${F}$ (·) denotes the perceptual feature extractor.

We also incorporated the MOCO-based similarity loss~\cite{chen2020mocov2}, which was shown to improve the quality of reconstructed images. This loss is defined as:
\begin{equation}
    \ell_{MOCO}( \styleimg, \texturecolor ) = 1 - dot((M(\styleimg), M(\texturecolor))
\end{equation}
where ${M}$ (·) denotes the pretrained MOCO~\cite{chen2020mocov2} network, and ${dot}$ (·, ·) is a dot product operation.

In summary, the total loss function is defined as:
\begin{equation}
    \ell(\styleimg, \texturecolor) = \lambda_{1}\ell_{2}(\styleimg, \texturecolor) + \lambda_{2}\ell_{LPIPS}(\styleimg, \texturecolor) + \lambda_{3}\ell_{MOCO}(\styleimg, \texturecolor)
    \label{eq:totalloss}
\end{equation}

where $\ell(\styleimg, \texturecolor)$ is the total loss of the style encoder and the structure encoder, and $\lambda_{1}$, $\lambda_{2}$, and $\lambda_{3}$ are the hyperparameters for weighting the loss functions.
\section{Experiment}
\label{sec:experiment}

\begin{table*}[!t]
    \caption{\tb{Quantitative evaluation.}  We show the quantitative comparision under conditional and unconditional setting on both Cars and Faces datasets.}
    \label{tab:baseline}
    \centering
    \resizebox{\linewidth}{!}{
        \begin{tabular}{cllllllllllll}
        \toprule
        & \multicolumn{6}{c}{Conditional} & \multicolumn{6}{c}{Unconditional} \\ 
        \cmidrule(lr){2-7} \cmidrule(lr){8-13}
        & \multicolumn{3}{c}{Cars} & \multicolumn{3}{c}{Faces} & \multicolumn{3}{c}{Cars} & \multicolumn{3}{c}{Faces} \\ 
        \cmidrule(lr){2-4} \cmidrule(lr){5-7} \cmidrule(lr){8-10} \cmidrule(lr){11-13}
        \multirow{-3}{*}{Method} & FID$\downarrow$ & KID$\downarrow$ & GIQA$\uparrow$ & FID$\downarrow$ & KID$\downarrow$ & GIQA$\uparrow$ & FID$\downarrow$ & KID$\downarrow$ & GIQA$\uparrow$ & FID$\downarrow$ & KID$\downarrow$ & GIQA$\uparrow$ \\
        \midrule
        Texture Fields~\cite{OechsleICCV2019} & 139.63 & 0.0968 & 3.8751 & 26.66 & \tb{0.0163} & \tb{6.9158} & 149.21 & 0.1156 & 3.8107 & 25.44 & 0.0344 & 6.6823\\
        LTG~\cite{rui2021LTG} & 149.82 & 0.1831 & 3.5379 & 24.31 & 0.0239 & 6.6326 & 87.06 & 0.0806 & 4.4394 & 20.08 & \tb{0.0203} & 6.7668\\
        Ours & \tb{39.41} & \tb{0.0646} & \tb{4.6071} & \tb{21.56} & 0.0259 & 6.8978 & \tb{66.73} & \tb{0.0652} & \tb{4.6276} & \tb{18.66} & 0.0212 & \tb{7.0326}\\ 
        \bottomrule
        \end{tabular}
    }
\end{table*}

This section presents results that validate the performance of {\networkName}. We begin by describing the experimental settings, which includes implementation details, the baselines, the dataset used, and the evaluation metrics (\secref{ExperimentalSetting}). Subsequently, we demonstrate the quantitative and qualitative results in comparison with two state-of-the-art baselines for texture generation tasks across two datasets under both style conditioned and unconditioned settings (\secref{Evaluation} and \secref{UnconditionEval}). Finally, we conducted two ablation studies to confirm the effectiveness of our design choices (\secref{Ablation}).


\subsection{Experimental Setting}
\label{sec:ExperimentalSetting}

\paragraph{Implementation Details.}

We implement our network using PyTorch and Python3.7 and conduct experiments on a single NVIDIA V100 with 32GB VRAM and training for 7 days.
We train both our style $\estyle$ and structure $\egeo$ encoders using the Adam~\cite{kingma2014adam} optimizer with a batch size of 8 and a learning rate of 0.0001 for 200k iterations.
The resolution of all training images are resized to $128 \times 128$.
We empirically set $\lambda_{1}=1.0$, $\lambda_{2}=0.8$, $\lambda_{3}=0.5$ in \eqnref{totalloss}, $n=6$, $N=6$ for the ShapeNet car dataset~\cite{shapenet2015} and $n=7$, $N=1$ the FFHQ human face dataset~\cite{Karras_2019_CVPR} in \secref{styleganEncoders} and \secref{textureMapping}.



\paragraph{Baselines.}
To assess the effectiveness of {\networkName}, we compare our method with two baselines. The first baseline is Texture Fields~\cite{OechsleICCV2019}, which generates textures of 3D shapes and represents them with an implicit field. For the condition setting, we retrain Texture Fields using both real-world and synthetic data to perform a fair comparison with our method. In contrast, our method is trained solely on real-world datasets to produce photo-realistic results. For the unconditional setting, we fine-tuned the Texture Fields pretrained weights using real-world images.

The second baseline we consider is LTG~\cite{rui2021LTG}, which learns texture generation in UV space with multiple "constant starting vectors" and discriminators. For the conditional setting, we employ pSp~\cite{richardson2021encoding} as the style encoder and use the encoded style code as input for LTG. For the unconditional setting, we retrain LTG on our training dataset using the official configuration. Both the conditional and unconditional settings are tested on the same data split to ensure fairness in the comparison.


Although we attempted to compare our method with Texturify~\cite{siddiqui2022texturify}, which generates realistic textures directly on the surface of 3D shapes by taking the 3D shape and style latent code as input, we encountered some challenges and the code of Texturify~\cite{siddiqui2022texturify} was not available for our comparison experiment.

\paragraph{Datasets.}
We evaluate our method under two datasets: CompCars dataset~\cite{yang2015compcars} (16,000/2000/2000 for train/validation/test split) and FFHQ~\cite{Karras_2019_CVPR} dataset (65,747/64/3000 for train/validation/test split). For more details on two datasets, please refer to our supplemental material.

\begin{figure*}[t]
    \centering
    \includegraphics[width=\textwidth]{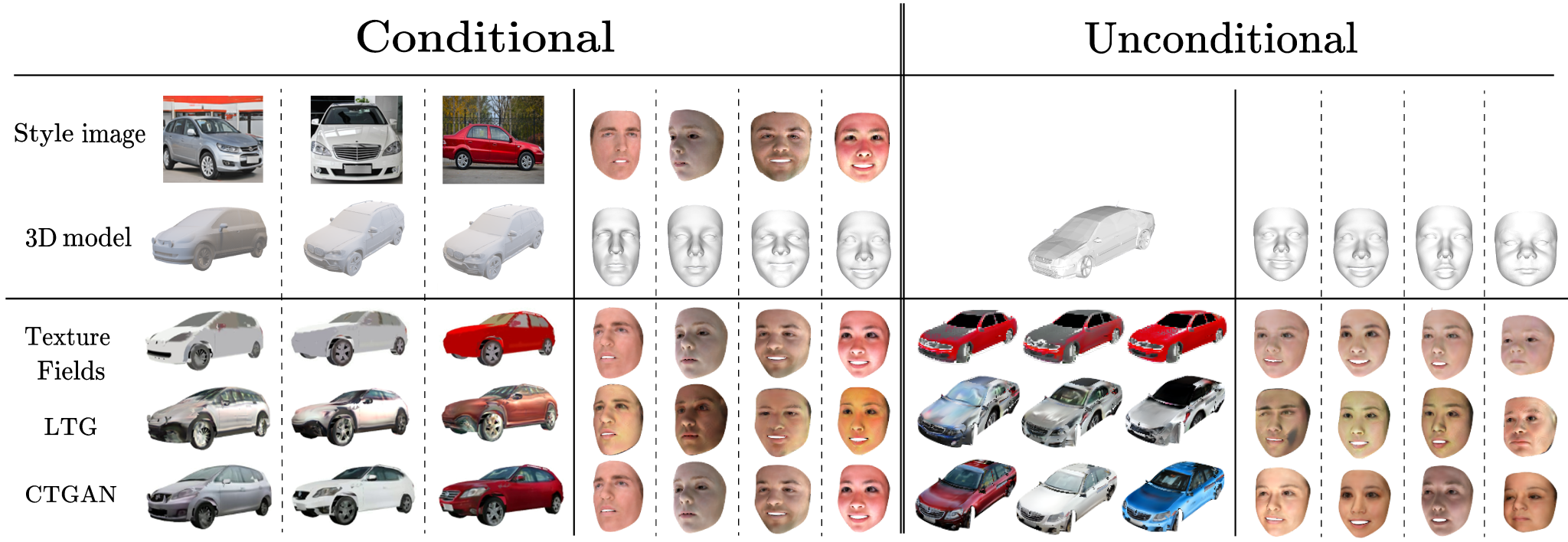}

    \caption{\tb{Qualitative comparison on texture generation.} First row: the input style images (only for the conditional part) and the input 3D models. Bottom 3 rows: the generated 3D textured models for each method using the input data from the first row.
    Our method produces superior results in generating texture maps that are more similar to style images and more view-consistent.}
    \label{fig:qualitative_result_1}
\end{figure*}

\paragraph{Metrics.}
To measure the quantitative performance, we follow the same evaluation process used in previous studies~\cite{OechsleICCV2019, rui2021LTG, siddiqui2022texturify}.
We employ three commonly used metrics to evaluate the realism and diversity of the generated results: Frechet Inception Distance (FID)~\cite{heusel2017fid}, Kernel Inception Distance (KID)~\cite{bińkowski2018demystifying}, and Generated Image Quality Assessment (GIQA)~\cite{gu2020giqa}.
As for the structure reconstruction accuracy in ablation study, we calculate the mIoU~\cite{long2015fully} and Pixel Accuracy~\cite{long2015fully} between the structure input and the predicted segmentation map of generated textures.

\subsection{Conditional Setting}
\label{sec:Evaluation}


In this experiment, we evaluate both the quantitative and qualitative performance of our model and compare it with other baselines on conditional setting (input with style images).
We report the quantitative result in the condition part of \tabref{baseline}. 
Our method achieves the best performance of all the metrics except the KID and GIQA on the faces dataset.
%
The qualitative comparisons are shown in the condition part of \figref{qualitative_result_1}.
As comparing the results of Texture Fields ~\cite{OechsleICCV2019} on cars dataset, our method could generate more realistic results and achieve better visual quality.
In contrast to LTG~\cite{rui2021LTG}, it is worth noting that our two-encoder architecture is capable of producing results which are perceptually aligned with the style input and view-consistent without any noticeable texture seam, and with much more fine-grained structural details. Please refer to our supplemental material for more results.


\subsection{Unconditional Setting}
\label{sec:UnconditionEval}
With regard to the unconditional setting, we evaluate both the quantitative and qualitative performance of our model and compare it with other baselines as well.
To demonstrate that our network can generate texture without a style image, we follow the same experimental settings as the conditional setting (\secref{Evaluation}) but replace the style latent code $\wstyle$ with the random vector sampled from a Gaussian distribution with 0 mean and 1.0 variance.
%
We report the quantitative result in the unconditional part of \tabref{baseline}. 
Our method achieves the best performance of all the metrics except the KID on faces dataset.
%
The qualitative comparisons of unconditional setting are illustrated in the unconditional part of \figref{qualitative_result_1}, where it shows that our method produces high-fidelity results with better structural details, while LTG suffers from distortion and view-inconsistent problems. 
Furthermore, our model can produce more diverse results than Texture Fields and LTG, which tend to generate results in specific colors.

\begin{table}[t]
\caption{\tb{Ablations.} We validate our design choices by comparing with several alternative options.}
\label{tab:ablation}
    \begin{subtable}[t]{0.16\textwidth}
        \centering
        \caption{Different Structure}
        \label{tab:two_style_code_encoder}
        \resizebox{\textwidth}{!}{
            \begin{tabular}{cc}
            \toprule
            Method & FID$\downarrow$ \\
            \midrule
            Silhouette & 42.86 \\ 
            Segmentation & \tb{39.41} \\ 
            \bottomrule
            \end{tabular}
        }
        \label{tab:two_style_code_encoder}
    \end{subtable}
    \begin{subtable}[t]{0.31\textwidth}
        \centering
        \caption{Different Structure Encoders}
        \resizebox{\textwidth}{!}{
            \begin{tabular}{clllllll}
            \toprule
            & &  \multicolumn{3}{c}{mIoU($\%$)$\uparrow$} & \multicolumn{3}{c}{Pixel Accuracy($\%$)$\uparrow$} \\ 
            \cmidrule(lr){3-5} \cmidrule(lr){6-8}
            \multirow{-2}{*}{Architecture} & \multirow{-2}{*}{FID$\downarrow$} & Front & Back & Side & Front & Back & Side\\
            \midrule
            pSp~\cite{richardson2021encoding} & 41.33 & 73.70 & 57.74 & 66.04 & 84.74 & 76.33 & 82.08 \\
            Ours & \tb{39.41} & \tb{74.28} & \tb{64.94} & \tb{69.31} & \tb{87.97} & \tb{82.58} & \tb{82.84} \\
            \bottomrule
            \end{tabular}
        }
        \label{tab:geometry_encoder}
    \end{subtable}
    \hfill
\end{table}

\subsection{Ablation Study}
\label{sec:Ablation}

Here, we present experiments to validate our model's architecture from different aspects.
%
We conduct two experiments to validate the necessity of the segmentation and the structure encoder $\egeo$ with coarse-to-fine architecture.
%
%
%
All the experiments are conducted on the cars dataset under the conditional setting.

\paragraph{The Effect of The Different Structure Inputs.}
\begin{figure}
    \centering
    \includegraphics[width=\linewidth]{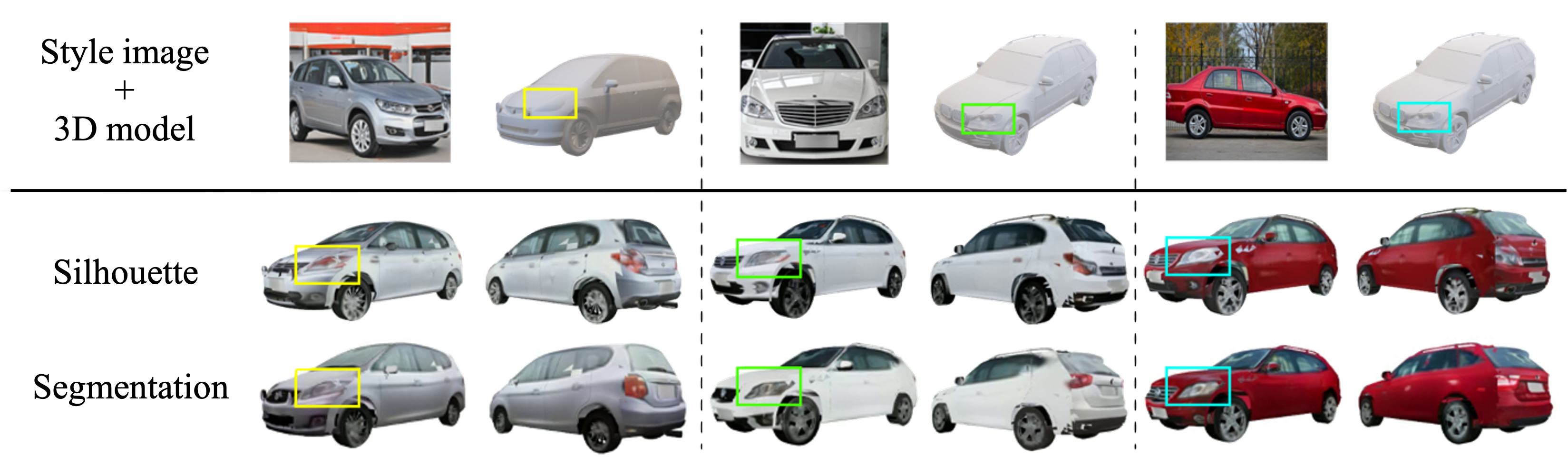}
    \caption{\tb{Qualitative comparison of different structure.}
    }
    \label{fig:two_encoder}
\end{figure}
We demonstrate the quantitative and qualitative differences between the different structure inputs.
Specifically, we replace our original structure input, which is segmentation maps, with the silhouettes.
As shown in \tabref{two_style_code_encoder}, we obtain the best performance with the segmentation maps as structure input (``Segmentation'' in \tabref{two_style_code_encoder}) on FID~\cite{heusel2017fid}.
The qualitative comparisons shown in \figref{two_encoder} indicate that we can produce the textures match the underlying 3D model ($2^{nd}$ and $3^{rd}$ rows in \figref{two_encoder}) well.

\paragraph{The Effect of The Different Structure Encoders.}

\begin{figure}[t]
    \centering
    \includegraphics[width=\linewidth]{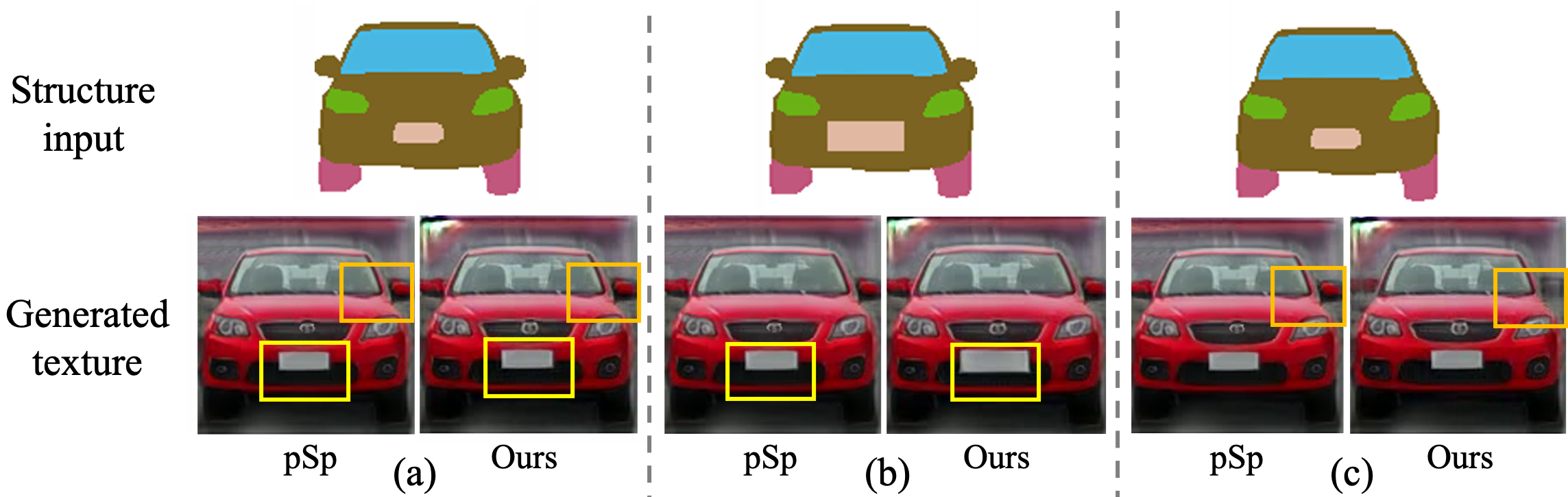}
    \caption{\tb{Qualitative comparisons on structure encoders.} 
    }
    \label{fig:geo_encoder_compare}
\end{figure}

We validate the necessity of the coarse-to-fine structure encoder.
In detail, we compare the quantitative and qualitative differences between the original pSp~\cite{richardson2021encoding} structure encoder and our coarse-to-fine structure encoder.
As shown in \tabref{geometry_encoder}, our method outperforms pSp~\cite{richardson2021encoding} in all metrics.
The qualitative comparisons shown in \figref{geo_encoder_compare} indicate that our coarse-to-fine structure encoder is more reflective of subtle changes \eg the different size of license plate between the a and b columns, and the absence of side mirror between a and c columns.


\section{Conclusions}
\label{sec:conclusion}
\begin{figure}
    \centering
    \includegraphics[width=.9\linewidth]{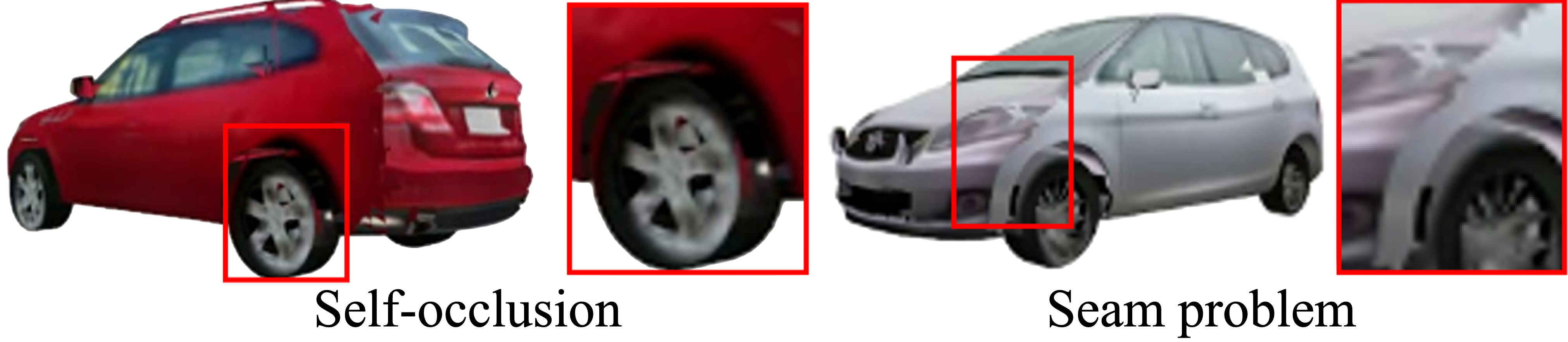}
    \caption{\tb{Limitation.} Our model produce the results struggling with handling self-occlusion geometry \emph{(left)} and seam problem \emph{(right)}.}
    \label{fig:limitation}
\end{figure}
We propose a novel conditional texture generator conditioned on the structural input and style image and adopting the StyleGAN2-ADA~\cite{Karras2020ada} as the backbone.
Our generated textures are high quality and can be well-aligned with the style images.
We propose two encoders to encode the structural input and style image separately, generating the texture with view consistency.
Our method outperforms existing methods both quantitatively and visually.

\heading{Limitations.}
View-based texture projection simplifies our architecture to work with StyleGAN2-ADA~\cite{Karras2020ada}, but it struggles with complex geometry or self-occlusion problems (left of \figref{limitation}) or seams at edges of different views (right of \figref{limitation}). 
Nevertheless, our model produces realistic and view consistent results.

\bibliographystyle{ACM-Reference-Format}
\bibliography{main}










\end{document}


\title{\networkName: Semantic-guided Conditional Texture Generator for 3D Shapes}

\author{First Author\\
Institution1\\
Institution1 address\\
{\tt\small firstauthor@i1.org}
\and
Second Author\\
Institution2\\
First line of institution2 address\\
{\tt\small secondauthor@i2.org}
}

\maketitle


\ificcvfinal\thispagestyle{empty}\fi

\input{sup_tex/results}

{\small
\bibliographystyle{ieee_fullname}
\bibliography{egbib}
}